\documentclass[letterpaper, 10pt, conference]{ieeeconf}  

\IEEEoverridecommandlockouts                              

\usepackage[utf8]{inputenc}
\usepackage{cite}
\usepackage{graphics} 
\usepackage{epsfig} 
\usepackage{epstopdf}
\usepackage{mathrsfs}
\usepackage{times} 
\usepackage{amsmath} 
\usepackage{amssymb}  
\usepackage{hyperref}
\usepackage{epstopdf}
\usepackage{subfigure}
\usepackage{epic}
\usepackage{bm}
\usepackage{tikz}
\usepackage{mdframed}
\usetikzlibrary{arrows,shapes,snakes,automata,backgrounds,petri}
\usepackage{algorithm}
\usepackage{algpseudocode}
\usepackage{url}
\usepackage{booktabs, caption, makecell}

\usepackage{threeparttable}

\usepackage{graphicx} 
\usepackage{color} 
\newcommand{\startpara}[1]{{\vskip1pt\noindent{\bf #1.}}}

\floatname{algorithm}{Algorithm}

\newtheorem{definition}{Definition}
\newtheorem{theorem}{Theorem}

\newtheorem{problem}{Problem}

\newtheorem{remark}{Remark}

\newtheorem{proposition}{Proposition}
\newtheorem{example}{Example}

\newmdenv[linecolor=black!20,backgroundcolor=gray!5]{examplebox}

\title{\LARGE \bf Trust-Aware Motion Planning for Human-Robot Collaboration under Distribution Temporal Logic Specifications}

\author{Pian Yu$^{1}$, Shuyang Dong$^{2}$, Shili Sheng$^{2}$, Lu Feng$^{2}$, and Marta Kwiatkowska$^{1}$
\thanks{*This work was supported in part by the ERC ADG FUN2MODEL (Grant agreement ID: 834115), NSF grant CCF-1942836, and AFOSR grant FA9550-21-1-0164. }
\thanks{$^{1}$Pian Yu and Marta Kwiatkowska are with the Department of Computer Science, University of Oxford,
        Oxford, United Kingdom
        {\tt\small {pian.yu, marta.kwiatkowska}@cs.ox.ac.uk}}%
\thanks{$^{2}$Shuyang Dong, Shili Sheng, and Lu Feng are with School of Engineering, University of Virginia, United States
        {\tt\small {sd3mn, ss7dr, lu.feng}@virginia.edu}}%
}

\begin{document}

\maketitle
\thispagestyle{empty}
\pagestyle{empty}

\begin{abstract}

Recent work has considered trust-aware decision making for human-robot collaboration (HRC) with a focus on model learning. In this paper, we are interested in enabling the HRC system to complete complex tasks specified using temporal logic that involve human trust. Since human trust in robots is not observable, we
adopt the widely used partially observable Markov decision process (POMDP) framework for modelling the interactions between humans and robots. To specify the desired behaviour, we propose to use syntactically co-safe linear distribution temporal logic (scLDTL), a logic that is defined over predicates of states as well as belief states of partially observable systems. The incorporation of belief predicates in scLDTL enhances its expressiveness while simultaneously introducing added complexity. This also presents a new challenge as the belief predicates must be evaluated over the continuous (infinite) belief space. To address this challenge, 
we present an algorithm for solving the optimal policy synthesis problem. First, we enhance the belief MDP (derived by reformulating the POMDP) with a probabilistic labelling function. Then a product belief MDP is constructed between the probabilistically labelled belief MDP and the automaton translation of the scLDTL formula. Finally, we show that the optimal policy can be obtained by leveraging existing point-based value iteration algorithms with essential modifications. Human subject experiments with 21 participants on a driving simulator demonstrate the effectiveness of the proposed approach.
\end{abstract}

\section{Introduction}

Autonomous robots are rapidly evolving into an essential component of our society, to mention home assistive robots \cite{pineau2003towards} and automated vehicles (AV) \cite{schwarting2018planning}. Despite
significant advances made in automation in recent years, attaining full autonomy, which would enable robots to successfully deal with complicated and unpredictable events or situations, remains stubbornly out of reach.  
For instance, the AV industry has had to reset expectations as it shifts its focus from level 5 to 4 autonomy \cite{anderson2020surprise}. In many applications where robots work with or alongside humans, it is customary to have robots that are operated or supervised by a human operator. Collaborative human-robot partnerships often hinge on the foundation of trust. Therefore, recognizing human trust and incorporating it into the decision-making process is essential for achieving the full potential of human-robot interactive systems.

The subject of trust 
has been actively studied in multiple  contexts such as psychology \cite{dunning2011understanding} and automation \cite{lee2004trust}. It is a multifaceted concept that can be influenced by a large number of factors. In the context of human-robot collaboration (HRC), studies have shown that the level of human trust in robots 
evolves over their interaction,
affected by factors such as the automation’s reliability, predictability, and transparency \cite{schaefer2016meta}. 
While earlier work has focused on studying the measurement \cite{freedy2007measurement}, modelling \cite{xu2015optimo}, and calibration \cite{wang2016trust} of human trust in robots, recent work has gravitated towards devising strategies that enable robots to proactively infer and influence the human collaborator’s trust \cite{chen2018planning, sheng2021trust}. 

Various methods exist for modeling the interaction between humans and robots. Among these, the game-theoretic approaches \cite{sadigh2016planning,huang2019reasoning} and the partially observable Markov decision process (POMDP) framework \cite{broz2008planning,lauri2022partially} have been extensively explored. Since trust is not fully observable, in this work we adopt the POMDP, where human trust can be modelled as a hidden variable. 
While the POMDP formulation allows the robot to act according to its beliefs about the human collaborator's trust based on observations, finding solutions to POMDPs of a realistic size is computationally challenging and existing work often relies on approximation algorithms \cite{silver2010monte,smith2012point,kurniawati2008sarsop}. Due to the inherent complexity of solving a POMDP, prior work devoted to trust-based decision making for HRC often  focused on relatively simple specifications (e.g., accumulated reward maximisation) \cite{chen2020trust, sheng2022planning}. Moreover, in all these studies, trust was treated as an implicit factor that impacts the performance of collaboration. None of these works has considered trust as part of the specification, where explicit requirements can be imposed. Real-world case studies have shown that an inappropriate
level of trust may result in the misuse or disuse of automation \cite{lee2004trust}. Therefore, in practical scenarios, it might be advantageous to stipulate conditions such as ``the trust level must not fall below a certain threshold" and ``the trust level at a particular juncture must surpass a certain threshold".

In this work, we investigate trust-aware motion planning for HRC with complex temporal logic specifications applied to both the state of the robot and the  trust (belief) of human. The trust-based human-robot interaction is modelled by a trust POMDP and syntactically co-safe linear distribution temporal logic (scLDTL) \cite{jones2013distribution} is  utilised to specify the desired behaviour of the system. 
Syntactically co-safe linear temporal logic (scLTL) is a commonly employed logic in robotics to specify the intended behaviour of a robot \cite{ayala2013temporal,cho2017cost}. In \cite{jones2013distribution}, scLDTL was introduced as an extension of scLTL to leverage the richness of information contained within belief states of partially observable systems.  It was shown that scLDTL is capable of expressing properties involving uncertainty and likelihood that cannot be described by existing logic. Nevertheless, the increased complexity introduced by the inclusion of belief predicates in scLDTL, which must be evaluated over the continuous (infinite) belief space, renders verification and synthesis from scLDTL a more demanding task. In \cite{jones2013distribution}, a feasibility checking algorithm was proposed for POMDPs with scLDTL specifications. However, to the best of our knowledge, the more challenging synthesis problem remains unresolved.
Our contributions are summarised as follows: i) We demonstrate the  suitability of scLDTL for specifying the desired behaviour of trust-aware HRC systems that involve requirements in the robot workspace as well as the trust (belief) space. ii) We propose an efficient algorithm to solve the scLDTL optimal policy synthesis for trust POMDPs, which overcomes the aforementioned complexity of scLDTL specifications. iii) We design and conduct human subject experiments with 21 participants on a driving
simulator to evaluate the proposed approach, with encouraging results.



\section{Motivating example}

\begin{figure}[t]
\centering
\subfigure[]{
\includegraphics[width=0.35\textwidth]{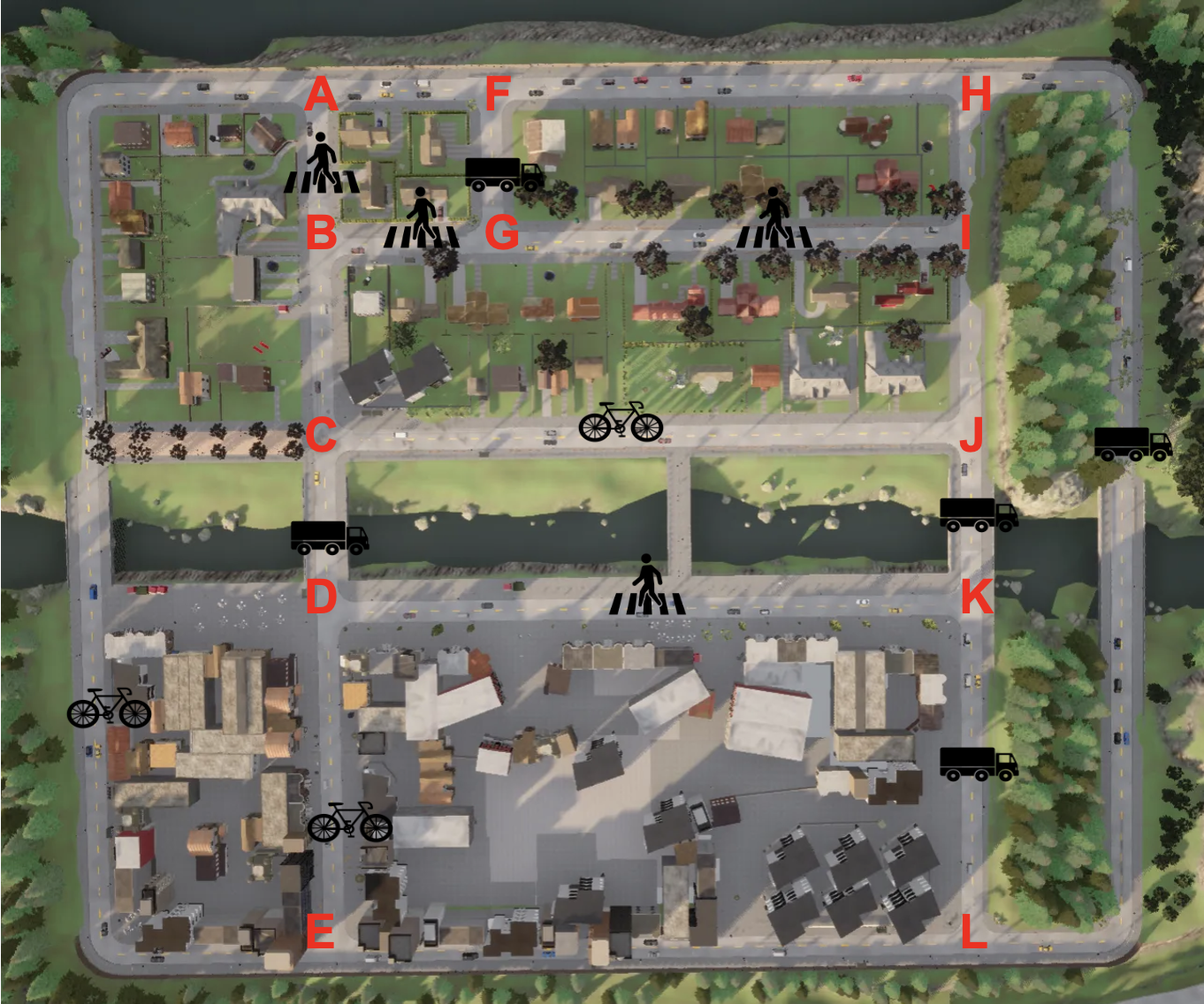}}\\
\subfigure[]{	\includegraphics[width=0.45\textwidth]{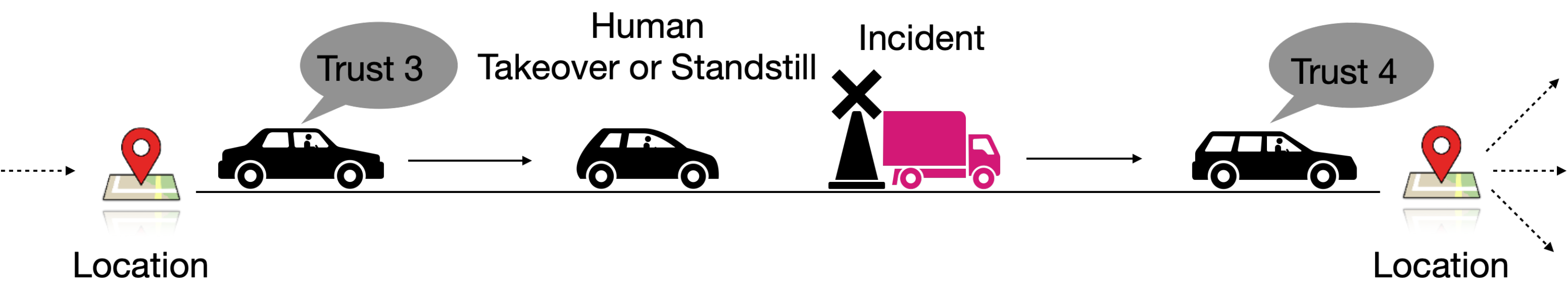}}
\caption{Route planning for AVs: a) A map with three types of road incidents (pedestrian, obstacle, and oncoming truck). b) A schematic view of the decision making process.}	\label{Fig:routeplanning}
\end{figure}

We describe a route planning problem for AVs. A human is driving an AV in a town, whose map is shown in Fig. \ref{Fig:routeplanning}(a). Within the town, we consider that there are 3 types of typical incidents that may occur on the road: (1) a pedestrian crossing the road, (2) an obstacle (e.g., a broken bicycle) ahead of the lane, and (3) an oncoming truck in the neighbouring lane. For simplicity, here we assume that there is at most one incident at a time for each road segment. 

Fig. \ref{Fig:routeplanning}(b) shows a schematic view of the AV traveling from one location to another. Imagine
the AV is approaching an incident on a road segment while in autopilot mode. For safety considerations, the driver might choose to take over control of the AV and switch to manual driving. The level of trust that the driver has in the AV's ability to handle various types of incidents can influence their takeover decision; a driver with lower trust is more inclined to do so. Furthermore, the driver's level of trust changes over time and depends on the takeover decision and the vehicle's ability to handle an incident. In our previous work \cite{sheng2022planning}, human subject experiments have shown that, by proactively inferring human trust and taking it into account during decision making, the AV can achieve higher cumulative rewards. 

The research focus of \cite{sheng2022planning} was on optimal route planning (e.g., navigating from one location to the other) for AVs. In contrast, in this work we are interested in trust-aware HRC in a boarder context. Our goal is to develop a trust-aware motion planning approach for HRC systems, which is capable of completing complex tasks specified in temporal logic 
that involve requirements on human trust levels.

\section{Preliminaries}

Before formulating our problem, we provide preliminary background on POMDPs and scLDTL.

\subsection{Partially observable Markov decision processes}

This section introduces POMDPs, which are well suited to the modelling of HRC systems under investigation. The human internal states (e.g., trust), which are not fully observable to robots, can be modelled as hidden states in POMDPs. 
In order to accurately represent the interactions between humans and robots, modifications to the conventional definition of a POMDP \cite{kaelbling1995partially} are incorporated.

\begin{definition}[POMDPs]\label{POMDP}
A POMDP is defined as a tuple $\mathcal{M}=(S, A, O, \mathcal{Z}, \mathcal{T})$, where
\begin{itemize}
    \item $S$ is a finite set of states;
    \item $A$ is a finite set of actions;
    \item $O$ is a finite set of observations;
    \item $\mathcal{Z}: S \times A \times O\to [0, 1]$ is the probabilistic observation function, which gives the probability of observing $o$ after taking action $a$ in state $s$, i.e., $\mathcal{Z}(s, a, o)=p(o|s, a)$;
    \item $\mathcal{T}: S\times A \times O\times S 
    \to 
    [0, 1]$ is the probabilistic transition function, which gives the probability that the state has value $s'$, after taking action $a$ and receiving observations $o$ in state $s$, i.e., $\mathcal{T}(s, a, o, s')=p(s'|s, a, o)$.
\end{itemize}
\end{definition}

Firstly, we note that, in Definition \ref{POMDP}, the probability of receiving observation $o\in O$ is determined by the previous state $s$ (instead of the resulting state $s'$) and the action $a$ that was just taken. Secondly, the transition function $\mathcal{T}$ is dependent on the observations. For the purpose of this work, a reward function is redundant and has been omitted.

Since a POMDP state is partially observable, we rely on the concept of a \emph{belief state}\footnote{A belief state is a probability distribution over all possible states in the POMDP. It represents the agent's subjective probability distribution of being in each state given its past observations and actions.}. Let $\mathcal{B}$ be the belief space of $S$.
A POMDP policy $\pi: \mathcal{B} \to A$ maps a belief state $b\in \mathcal{B}$, which
is a probability distribution over $S$, to a prescribed action $a\in A$. 
Given a policy $\pi$, the control of the agent’s actions is performed
online. First, the agent takes an action $a=\pi(b)$ according to the given
policy $\pi$ and the current belief
is $b$. Second, after taking an action
$a$ and receiving an observation $o$, the agent updates its
belief:
\begin{equation}\label{belief_update}
    b'(s')=\sum_{s\in S}b(s){\sum_{o\in \mathbb{F}_O(s, a)}\mathcal{Z}(s, a, o)\cdot \mathcal{T}(s, a, o, s')},
\end{equation}
where $\mathbb{F}_O(s, a)=\{o\in O| \mathcal{Z}(s, a, o)>0\}$. The process then repeats. An interesting property to note about the POMDP described in Definition \ref{POMDP} is that the belief update (\ref{belief_update}) is linear\footnote{The belief update of a conventional POMDP is often represented using the Bayes' filter.}. An \emph{execution} $\bm{\rho}$ of a POMDP is a possibly infinite alternating sequence of belief states, actions, and observations, i.e., $\bm{\rho}=b_0a_0o_0b_1a_1o_1\cdots$.

\subsection{Syntactically Co-Safe Linear Distribution Temporal Logic}

This section introduces scLDTL for concisely specifying the desired behaviour of the HRC systems. It will become clear later that scLDTL is capable of expressing requirements in both the robot workspace and the trust (belief) space.

scLDTL consists of two types of predicates: i) state predicates $\nu\in 2^S$ and ii) belief predicates $\mu$, which are obtained after evaluation of a predicate function $g_\mu: \mathcal{B}\to \mathbb{R}$ on the belief space $\mathcal{B}$ as follows
\begin{equation*}
   \mu:=\left\{\begin{aligned}
   \top, & \quad \text{if } \quad g_\mu(b)< 0 \\
   \bot, & \quad \text{if } \quad g_\mu(b)\ge 0.
   \end{aligned}\right.
\end{equation*}

An scLDTL formula is defined inductively
according to the following syntax \cite{jones2013distribution}:
\begin{equation}\label{LDTL}  \varphi::=\top|\nu|\mu|\neg\nu|\neg\mu|\varphi_1\wedge\varphi_2|\varphi_1\vee\varphi_2|\varphi_1 \mathsf{U}\varphi_2|\bigcirc \varphi|\lozenge \varphi,
\end{equation}
where $\nu\in 2^S$ is a set of states, $\mu$ is a belief predicate, $\neg$
 (negation), $\wedge$ (conjunction), and $\vee$ (disjunction) are 
logic connectives, and $\mathsf{U}$ (until), $\bigcirc$ (next) and $\lozenge$ (eventually) are temporal operators. We omit the scLDTL semantics due to
page limit and refer the reader to \cite{jones2013distribution}.

Let ${\rm AP}$ be a set of state predicates and ${\rm BP}$ be a set of belief predicates. The satisfaction of an scLDTL formula  $\varphi$ over ${\rm AP}\cup {\rm BP}$ can be captured through a deterministic finite automaton (DFA) $\mathcal{A} = (Q, q_0, 2^{\rm AP\cup BP}, \delta, {\rm Acc})$, where 
\begin{itemize}
  \item $Q$ is a finite set of states,
  \item $q_{0}\in Q$ is the initial state,
  \item $\delta: Q\times 2^{\rm AP\cup BP} \to Q$ is the transition function, and
  \item ${\rm Acc}\subseteq Q$ is the set of accepting states.
\end{itemize}
A finite \emph{run} $q=q_0q_1\ldots q_k$ of $\mathcal{A}$ is called \emph{accepting} if $q_k\in {\rm Acc}$. 

Now we define the notion of probabilistic satisfaction with respect to an execution $\bm{\rho}$ of a POMDP.

\begin{definition}\label{def:LDTLsatisfaction}[scLDTL satisfaction with respect to a POMDP execution]
Given an execution $\bm{\rho}=b_0a_0o_0b_1a_1o_1\ldots$ of a POMDP $\mathcal{M}$, the probability that the execution $\bm{\rho}$ satisfies the scLDTL formula $\varphi$ is given by
\begin{equation*}
   {\rm Pr}_{\mathcal{M}}(\{\text{$s_0s_1\cdots$ such that $(s_0, b_0)(s_1, b_1)\cdots \models \varphi$}\}\mid \bm{\rho}).
\end{equation*}
For simplicity, it is denoted in shorthand as ${\rm Pr}_{\mathcal{M}}(\varphi \mid \bm{\rho}).$
\end{definition}

\section{POMDPs for HRC}

In this work, we consider a human and a robot working collaboratively in the workspace $X$.
The human (H) adopts a supervisory role and the robot (R) is charged 
with performing tasks. The human can intervene in the task execution due to, for instance, low trust.

\subsection{HRC modelling}

Within the workspace $X$, one can identify a set of incidents $I$, i.e., a set of events that can affect human trust and/or takeover decision. The likelihood of observing an incident $id\in I$ is determined by the current state and the action of the robot. 
Denote by $\Theta$ the state space of human trust in the robot, which is not observable by the robot. 


The human-robot interaction can be modelled as a POMDP $\mathcal{M}$ as per Definition \ref{POMDP}, where the state space $S$ of $\mathcal{M}$ is factored as the observable state space $X$ and the non-observable state space $\Theta$, i.e., $S=X\times \Theta$. Accordingly, the probabilistic transition function $\mathcal{T}$ is factored as the world state and the human trust probabilistic transition functions $\mathcal{T}_X$ and $\mathcal{T}_\Theta$, respectively. It has been shown in \cite{chen2020trust} that the human trust affects human behaviour (e.g., takeover decision) and the human trust is affected by factors such as the robot performance (i.e., success/fail in handling an incident). Therefore, the state evolution of the trust POMDP $\mathcal{M}$ is determined not only by the robot action $a^r$, but also posterior observations, including the incident $id$ encountered during execution, the human takeover decision $a^h$, and the robot performance $e^r$. 
Formally, the trust POMDP is specified as a tuple
$\mathcal{M}=(X, \Theta, A^r, O, \mathcal{Z}, \mathcal{T}_X, \mathcal{T}_{\Theta})$, where 
\begin{itemize}
    \item $A^r$ is the finite action space of the robot;
    \item $O = I \times A^h \times \mathcal{E}^r$ is the observation set, where 
    \begin{itemize}
        \item $I$ is the set of incidents;
        \item $A^h=\{\rm{tk, st}\}$ is the action space of the human, where ${\rm tk}$ and ${\rm st}$ stand for ``takeover" and ``standstill", respectively; similarly to \cite{chen2020trust}, we assume that the human first observes
    the robot’s action $a^r$ and then decides his or her own action $a^h$;
        \item $\mathcal{E}^r=\{\rm{succ, fail}\}$ represents the performance of the robot, where ${\rm succ}$ and ${\rm fail}$ stand for ``success" and ``failure", respectively.
    \end{itemize} 
    \item $\mathcal{Z}: X \times \Theta \times A^r \times O \to [0, 1]$ is the probabilistic observation function, which is given by $\mathcal{Z}(x, \theta, a^r, o)= p(id, a^h, e^r|x, \theta, a^r);$
\item $\mathcal{T}_X: X \times A^r \times O \times X \to [0, 1]$ is the world state probabilistic transition function, i.e., $\mathcal{T}_X(x, a^r, o, x')=p(x'|x, a^r, o);$

\item $\mathcal{T}_\Theta: \Theta \times O\times \Theta \to [0, 1]$ is the human trust probabilistic transition function, i.e., $\mathcal{T}_\Theta(\theta, o, \theta')=p(\theta'|\theta, o)$.
\end{itemize}

A graphical model of the trust POMDP is shown in Fig. 1. It contains two key components: (i) a
trust dynamics model, which captures the evolution of human
trust in the robot, and (ii) a human decision model, which
connects trust with human actions. 

Let $\mathcal{B}_\Theta$ be the belief space of $\Theta$. We associate with each value
$x\in X$ a belief space for $\theta$: $\mathcal{B}_\Theta(x):=\{(x, b_{\Theta})| b_\Theta\in \mathcal{B}_\Theta\}$. Then the belief space of the trust POMDP $\mathcal{M}$ can be defined as
\begin{equation}\label{belief}
    \mathcal{B}=\cup_{x\in X} \mathcal{B}_\Theta(x).
\end{equation}
The POMDP policy $\pi:\mathcal{B} \to A^r$ maps a belief $b \in \mathcal{B}$ to a prescribed robot action $a^r\in A^r(b)$, where $A^r(b)$ is the set of actions available at belief $b$.

\begin{figure}[t]
\centering	\includegraphics[width=0.25\textwidth]{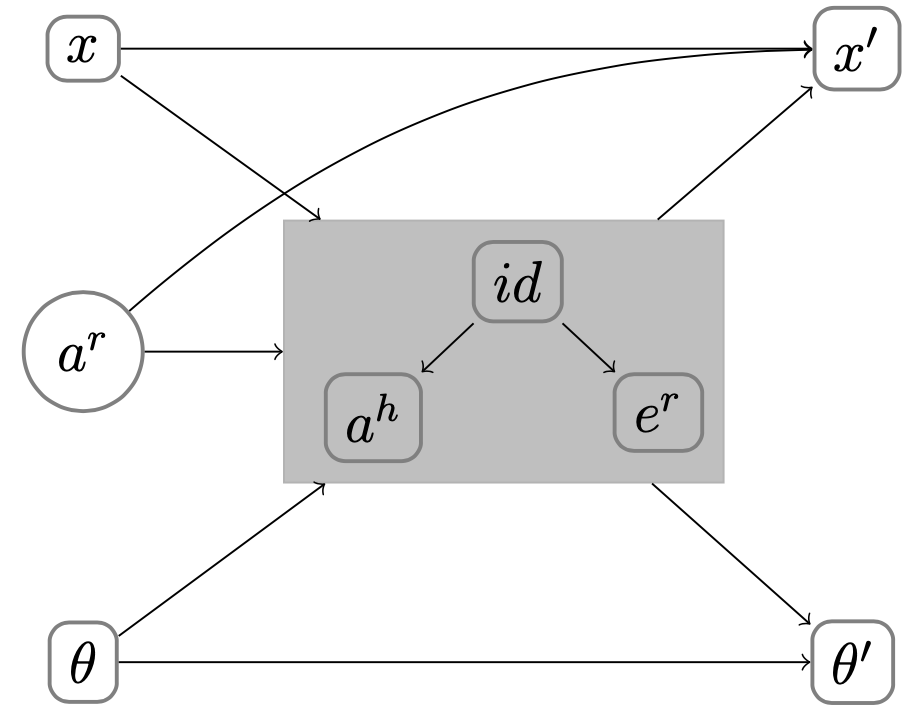}
	\caption{\footnotesize The graphical model for the trust POMDP $\mathcal{M}$. }
	\label{Fig:example}
	\vspace{-0.2cm}
\end{figure}

We illustrate our results with the motivating example.

\begin{example}[continued]
The town map shown in Fig. \ref{Fig:routeplanning}(a) has 12 road intersections $\{\rm{A, \cdots, L}\}$. Depending on the driving direction, each intersection can be factored into 3 different states (for instance, intersection $A$ contains states ${\rm EA}$, ${\rm BA}$, and ${\rm FA}$). We use Muir’s questionnaire \cite{muir1990} with a 7-point Likert scale as a human trust metric (i.e., trust ranges from 1 to 7). Therefore, one has that $X=\{{\rm EA}, {\rm BA}, {\rm FA} \cdots, {\rm EL}, {\rm KL}, {\rm HL}\}$, $\Theta=\{1, \cdots, 7\}$, and $I=\{\rm{`pedestrian', `obstacle', `truck'}\}$. The robot action is route choices and one can define an indicator function $\mathcal{I}$ for incidents. For instance, $\mathcal{I}({\rm EA}, {\rm AB}, {\rm `pedestrian'})=1$.
\end{example}

\subsection{Problem formulation}


In this work, we consider that the HRC system is required to complete complex tasks in the robot workspace $X$. Moreover, there are requirements on the human trust $\Theta$ and/or trust belief $b_\Theta$. We formulate the tasks in the workspace as well as the requirements on human trust and trust belief in the form of an scLDTL formula $\varphi$. 

\begin{example}[continued] 
  Consider now that the AV is required to complete the following tasks: i) visits the target intersections $G, J$ and $L$ (in this order) from the initial location ${\rm EA}$ (see Fig. \ref{Fig:routeplanning}(a)) and ii) the human trust level can not be too low (lower than 2) at all times, and iii) when the vehicle reaches the final intersection $L$, the likelihood that the human trust level is high (higher than or equal to 6) is no less than 0.5. In scLDTL, this specification can be written as 
  \begin{equation*}
      \varphi=(\neg \textsc{lowtrust})\mathsf{U}(\diamondsuit (\nu_1 \wedge \diamondsuit (\nu_2 \wedge \diamondsuit (\nu_3 \wedge \textsc{hightrust}))),
  \end{equation*}
where there are 3 state predicates $\nu_1 =\{{\rm BG}, {\rm FG}, {\rm IG}\}, \nu_2 =\{{\rm IJ}, {\rm CJ}, {\rm KJ}\}, \nu_3 =\{{\rm KL}, {\rm EL}, {\rm HL}\}$ and 2 belief predicates $\textsc{lowtrust}, \textsc{hightrust}$. The predicate functions are given by $g_{\textsc{lowtrust}} =1-A_1 b_\Theta, g_{\textsc{hightrust}} =0.5-A_2 b_\Theta$,
where $A_1= [1, 0, 0, 0, 0, 0, 0]$ encodes trust lower than 2 and $A_2= [0, 0, 0, 0, 0, 1, 1]$ encodes trust higher than or
equal to 6.
\end{example}

Given the initial belief state $b_0$ and a policy ${\pi}$, denote by ${\bm{\rho}}^{\pi}(b_0)$ 
the set of all possible executions generated by ${\pi}$. We consider the optimal policy synthesis problem for HRC under scLDTL specifications, i.e., find an optimal policy $\pi$ such that the probability of the set of all executions that satisfy an scLDTL formula $\varphi$ under $\pi$ is maximised. Mathematically,
this problem can be formulated as follows.

\begin{problem}\label{problem}
Given the trust POMDP ${\mathcal{M}}$ and the scLDTL specification $\varphi$, find a policy $\pi\in \Pi$ such that 
\begin{eqnarray} \label{Problem_formulation}
  &&\hspace{-0.5cm}\max_{\pi\in \Pi} \quad {\rm Pr}_{{\mathcal{M}}}^\pi(\varphi) \triangleq \sum_{\bm{\rho}\in \bm{\bm{\rho}}^{\pi}(b_0)} {\rm Pr}(\bm{\rho}|b_0, \pi){\rm Pr}_{{\mathcal{M}}}(\varphi\mid \bm{\rho}),
\end{eqnarray}
where $\Pi$ is the set of all policies for the trust POMDP ${\mathcal{M}}$ and ${\rm Pr}_{{\mathcal{M}}}(\varphi\mid \bm{\rho})$ is given in Definition \ref{def:LDTLsatisfaction}.
\end{problem}

\begin{remark}
    Point-based value iteration (PBVI) algorithm has been proposed for POMDPs under LTL specifications. In \cite{bouton2020point}, the atomic propositions of an LTL formula are evaluated on the state space of the POMDP, which is finite. Therefore, the construction of the product POMDP and the computation of maximal end components are similar to finite MDPs, for which the existing graph-based methods \cite{baier2008principles} can be utilised. In this work we consider scLDTL specifications, in which the belief predicates are evaluated over the belief space $\mathcal{B}$ of the trust POMDP ${\mathcal{M}}$, which is infinite. Therefore, the approach proposed in \cite{bouton2020point} is not applicable here. 
\end{remark}

\section{Proposed Approach}

This section presents our approach to solve the optimal policy synthesis problem (Problem 1), which falls outside the purview of existing policy synthesis algorithms designed for POMDPs. It is divided into two parts: 1) the construction of the product POMDP with the
DFA of the scLDTL formula $\varphi$ and 2) an algorithm to approximately compute a policy that maximises the probability of satisfying the given scLDTL formula $\varphi$.

\subsection{Product POMDP}

To begin with, we define the deterministic  state and belief predicate labelling functions $L_s$ and $L_b$ as
\begin{itemize}
    \item $L_s: X\times \Theta \to 2^{\rm AP}$, which contains the set of state predicates that can be true at state $(x, \theta)$;
    \item $L_b: X\times \mathcal{B}_\Theta \to 2^{\rm BP}$, which contains the set of belief predicates that can be true at belief state $(x, b_\Theta)$.
\end{itemize}
Then the associated probabilities are given by $p_{L_s}: X\times \Theta \times 2^{\rm AP} \to \{0, 1\}$ and  $p_{L_b}: X\times \mathcal{B}_\Theta \times 2^{\rm BP} \to \{0, 1\}$, respectively.

Now we propose to reformulate the trust POMDP $\mathcal{M}$ (equivalently) as a belief MDP and further expand it by including probabilistic labels, which gives a \emph{probabilistically labelled} belief MDP $\hat{\mathcal{M}}=(\mathcal{B}, A^r, O, \hat{\mathcal{Z}}, \hat{\mathcal{T}}_{\mathcal{B}}, \hat{L}, p_{\hat{L}})$, where $\mathcal{B}$ is defined in (\ref{belief}), $A^r, O$ are given in $\mathcal{M}$, and
   \begin{itemize}
      \item $\hat{\mathcal{Z}}: \mathcal{B} \times A^r \times O \to [0, 1]$ is the probabilistic observation function;
\item $\hat{\mathcal{T}}_{\mathcal{B}}=(\mathcal{T}_X, \mathcal{T}_{\mathcal{B}_\Theta})$ is the probabilistic transition function,  where $\mathcal{T}_X$ is defined in $\mathcal{M}$ and $\mathcal{T}_{\mathcal{B}_\Theta}: \mathcal{B}_\Theta \times O\times \mathcal{B}_\Theta \to [0, 1]$;
\item $\hat{L}: \mathcal{B}\to 2^{2^{\rm AP\cup BP}}$ is the belief state labelling function, where $\hat{L}(x, b_\Theta)$ contains the
set of state and belief predicate subsets that can be true at $(x, b_\Theta)$;
\item $p_{\hat{L}}: \mathcal{B}\times 2^{\rm AP\cup BP}\to [0, 1]$ specifies the associated probability. 
\end{itemize}
The probabilistic labelling function $p_{\hat{L}}$ provides a unified way of assigning belief states with both state and belief predicates in an scLDTL formula. For instance, given a belief state $(x, b_\Theta)$, a state predicate $\nu$, and a belief predicate $\mu$, one has that $p_{\hat L}(x, b_\Theta, \nu) = b_\Theta(\theta)p_{L_s}(x, \theta, \nu), \forall \theta\in \Theta$ and $p_{\hat L}(x, b_\Theta, \mu) = p_{L_b}(x, b_\Theta, \mu)$.

The product belief MDP $\mathcal{M}^{\times}$ is constructed between the probabilistically labelled belief MDP $\hat{\mathcal{M}}$ and the DFA $\mathcal{A} = (Q, q_0, 2^{\rm AP\cup BP}, \delta, {\rm Acc})$ of the scLDTL formula $\varphi$. 

\begin{definition}[Product belief MDP]\label{productautomaton}
Denote by $\mathcal{M}^{\times}$ the product $\hat{\mathcal{M}} \times \mathcal{A}$ as a tuple $\mathcal{M}^{\times} =(S^\times, A^r, O, {\mathcal{Z}}^\times, \mathcal{T}^\times, {\rm Acc}^\times)$, where 
\begin{equation*}
    S^\times=\mathcal{B}\times 2^{\rm AP \times BP} \times Q
\end{equation*}
is so that $(x, b_\Theta, l, q)\in S^\times, \forall (x, b_\Theta)\in \mathcal{B}, \forall l\in \hat{L}(x, b_\Theta)$, and $\forall q\in Q$, $\mathcal{Z}^\times: S^\times  \times A^r \times O \to [0, 1]$ is the probabilistic observation function, and ${\rm Acc}^\times =\{(x, b_\Theta, l, q)\in S^\times |q\in {\rm Acc}\}$. The probabilistic transition function $\mathcal{T}^{\times}: S^\times  \times A^r \times O \times S^\times \to [0, 1]$ is defined as
    \begin{equation*}
    \begin{aligned}
         &\mathcal{T}^{\times}((x, b_\Theta, l, q), a^r, o, (x', b'_\Theta, l', q'))=\\
        &\hspace{-0.3cm}\begin{cases}
             \hat{\mathcal{T}}_{\mathcal{B}}((x, b_\Theta), a^r, o, (x', b'_\Theta))\cdot p_{\hat L}(x', b'_\Theta, l'), & \text{if $q'\in \delta(q, l),$}\\
             0, & \text{otherwise};
         \end{cases}
    \end{aligned}       
    \end{equation*}
where $\hat{\mathcal{Z}}, \hat{\mathcal{T}}_{\mathcal{B}},$ and $p_{\hat L}$ are defined in $\hat{\mathcal{M}}$.
\end{definition}

Let $\Pi^\times$ be the set of all policies for the product belief MDP $\mathcal{M}^\times$. 
The set of \emph{accepting states} of $\mathcal{M}^\times$ is given by ${\rm Acc}^\times$.
We have the following result.

\begin{theorem}\label{prop1}
    Given the trust POMDP ${\mathcal{M}}$ and the scLDTL formula $\varphi$, the maximal probability of satisfying $\varphi$ is:
    \begin{equation*}
    \begin{aligned}
        &\max_{\pi\in \Pi}\{{\rm Pr}_{{\mathcal{M}}}^\pi(\varphi)\}
        = \max_{{\hat{\pi}}\in {\Pi^\times}} \{{\rm Pr}_{{\mathcal{M}}^\times}^{{\hat{\pi}}}( \diamondsuit {\rm Acc}^\times)\}.
    \end{aligned}       
    \end{equation*}
\end{theorem}

Theorem \ref{prop1} shows that, with the set of accepting states ${\rm Acc}^\times$, the original optimal policy synthesis
problem reduces to a reachability problem.

\subsection{Optimal policy synthesis}

PBVI algorithms have been widely used for solving POMDP synthesis problems  \cite{silver2010monte, smith2012point, kurniawati2008sarsop}. They often offer convergence guarantees specified
as upper and lower bounds on the value function. However, these PBVI algorithms are not directly applicable for solving our problem (Problem \ref{problem}). This is because solving a reachability problem for POMDPs necessitates the presence of a clearly defined reward function, which assigns value 1 to states in the goal set and 0 otherwise. However, in our case, capturing the satisfaction of an scLDTL specification through a state-based reward function is not feasible due to the presence of belief predicates.

In the following, we show that, with essential modifications, the existing PBVI algorithms can be leveraged for solving Problem \ref{problem} 
with the set of accepting states ${\rm Acc}^\times$.

Given a state $s$ of the product belief MDP $\mathcal{M}^\times$, we first define a value function $V: S^\times\to \mathbb{R}_{\ge 0}$ as
\begin{equation*}
    V(s)=\max_{{\pi}\in {\Pi^\times}} \{{\rm Pr}_{{\mathcal{M}}^\times}^{{\pi}}(\diamondsuit {\rm Acc}^\times)\},
\end{equation*}
which represents the maximal probability of reaching ${\rm Acc}^{\times}$ from initial state $s$. Then one can get that $V(s)=1, \forall s\in {\rm Acc}^\times$. For $s\notin {\rm Acc}^\times$, 
we further define the dynamic programming operator $T$ as
\begin{equation*}
\begin{aligned}
    &T({V})(s)\\
    =&\max\limits_{a^r\in A^r} \left\{\sum\limits_{o\in \mathbb{F}_O(s, a^r)}{\mathcal{Z}}^\times(s, a^r, o)\sum\limits_{s'\in S^\times}\mathcal{T}^\times(s, a^r, o, s'){V}(s')\right\}.
    \end{aligned}
\end{equation*}

Before running a PBVI algorithm, 
first we initialize the upper- and lower-bounds of the value function $V$ as follows: 
\begin{equation}\label{initial_upper}
    \overline{V}^0(s)=1, \forall s\in S^{\times}, \quad \underline{V}^0(s)=\begin{cases}
    1, & \text{if $s\in {\rm Acc}^\times,$}\\
    0, & \text{otherwise.}
    \end{cases}
\end{equation}
Then a
precision parameter $\tau$ is provided that controls the tightness
of the convergence (for example, by controlling the depth of the tree in
SARSOP \cite{kurniawati2008sarsop}), which yields $|\overline V(s_0)-\underline{V}(s_0)|\le \tau,$
where $s_0=(b_0, q_0)$ is the initial state of the product belief MDP $\mathcal{M}^\times$.

Denote by ${\rm Pr}_{{\mathcal{M}}}^{\max}(\varphi):=\max_{\pi\in \Pi}\{{\rm Pr}_{{\mathcal{M}}}^\pi(\varphi)\}$ the maximal probability of satisfying the scLDTL formula $\varphi$. We have the following result.

\begin{theorem}\label{thm2}
Let $\overline V(s_0)$ and $\underline{V}(s_0)$ be the upper- and lower-bounds of $V(s_0)$ obtained using PBVI with the initialization function (\ref{initial_upper}). One has that
\begin{equation*}
    \underline{V}(s_0)\le {\rm Pr}_{{\mathcal{M}}}^{\max}(\varphi)\le \overline{V}(s_0).
\end{equation*}
\end{theorem}

Finally, the optimal policy $\hat{\pi}^*$ for state $s\in S^\times$ can be derived using the value function.


\section{Implementation}

We have implemented the proposed approach to obtain the optimal policy for each scLDTL specification under consideration for the motivating example. To construct the trust POMDP, we utilise the trust dynamics model and the human takeover decision model, which were derived through an online user study involving 100 anonymous participants on Amazon
Mechanical Turk (AMT) platform \cite{sheng2022planning}.
Then two scLDTL specifications $\varphi_1$ and $\varphi_2$ are considered for the AV, 
\begin{align*}
    &&\hspace{-0.5cm}\varphi_1=(\neg \textsc{lowtrust})\mathsf{U}(\diamondsuit (\nu_1 \wedge \diamondsuit (\nu_2 \wedge \diamondsuit (\nu_3 \wedge \textsc{hightrust}))), \\ &&\hspace{-0.5cm}\varphi_2=(\neg \textsc{lowtrust})\mathsf{U}(\diamondsuit (\nu_4 \wedge \diamondsuit (\nu_5 \wedge \diamondsuit (\nu_6 \wedge \textsc{hightrust}))),
\end{align*}
 where $\nu_1, \nu_2, \nu_3, \textsc{lowtrust}, \textsc{hightrust}$ are defined in the Example and $\nu_4 =\{{\rm AB}, {\rm GB}, {\rm CB}\}, \nu_5 =\{{\rm DK}, {\rm JK}, {\rm LK}\}, \nu_6 =\{{\rm AE}, {\rm DE}, {\rm LE}\}$. The DFA for each scLDTL specification $\varphi_i, i\in \{1, 2\}$  is derived using \cite{klein2007ltl2dstar}. Then the corresponding product belief MDP $\mathcal{M}_i^\times$ is constructed with the DFA of $\varphi_i$. Finally, the upper- and lower-bounds of the value function $V_i$ are computed using the POMDP
toolkit “pomdp\_py” \cite{zheng2020pomdp_py} (with essential modifications describled in Section V.B). The precision parameter is set as $\tau = 0.01$. 
All simulations are carried out on a Macbook Pro (2.6 GHz 6-Core Intel Core i7 and 16 GB of RAM).

\section{Driving Simulator Study}
We evaluate the effectiveness of obtained policies via a driving simulator study.~\footnote{This study was approved by the University of Virginia Institutional Review Boards under IRB-SBS protocol \#6045. }

\subsection{Study design}
\begin{figure}[t]
\centering	\includegraphics[width=0.4\textwidth]{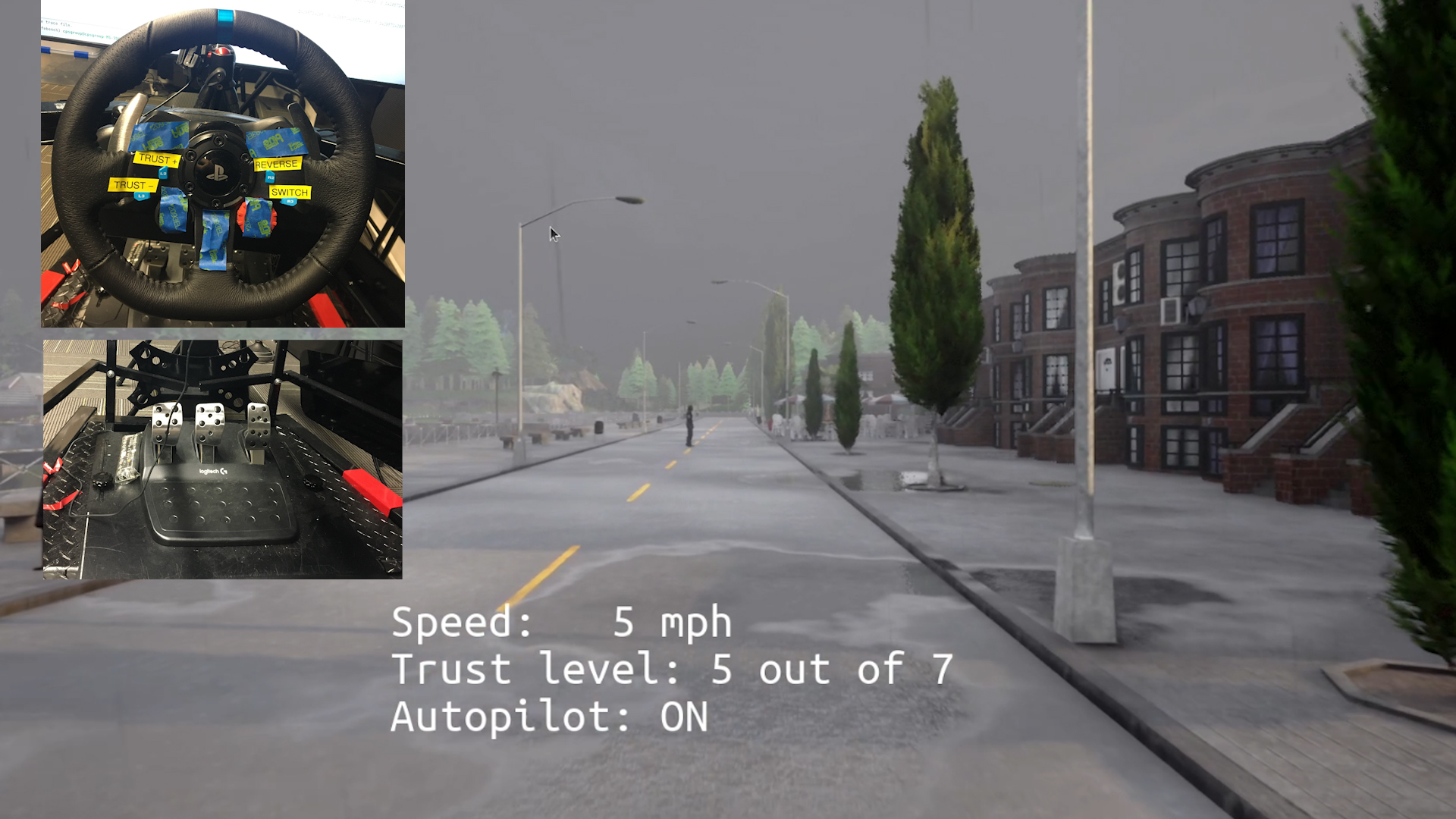}
	\caption{\footnotesize Driving simulator setup. }
	\label{Fig:simulator}
	\vspace{-0.2cm}
\end{figure}

\startpara{Apparatus} 
The study was conducted in a fixed-based driving simulator from SimXperience, consisting of a 55-inch display, a racing car seat, a Logitech G29 steering wheel, and sport pedals, see Fig. \ref{Fig:simulator}. In our study, four buttons on the steering wheel were programmed to let the drivers increase/decrease trust, switch driving mode between manual and autopilot, switch gear between drive and reverse.

\startpara{Driving Scenario} The experiments were run on a machine with 3.5GHz CPU, NVIDIA GeForce RTX 3080 Ti, 62GB memory, and Ubuntu 20.04.6 LTS operating system. The virtual driving environment was created using CARLA 0.9.13. An autopilot controller was programmed for several driving tasks such as lane keeping, taking turns at intersections, and handling incidents.



\startpara{Subject allocation}
We recruited 21 participants from the university community. All participants had a valid driver's license and regular or corrected-to-normal vision.
Each participant was compensated with a \$10 gift card. 
We adopted a within-subject study design: each participant took 4 unique routes, i.e., trust-aware and trust-free routes for both scLDTL specifications. The start and destination is either from A to L ($\varphi_1$) or from H to E ($\varphi_2$).
The route is either the trust-aware route (obtained using the computed optimal policy) or the trust-free route (which is the shortest-distance route from the start to the destination). However, if the vehicle has not reached the destination after travelling 20 intersections, it reschedules a shortest-distance route. The order of trials are randomized and counter-balanced.

\subsection{Study Procedure}
Upon arrival, a participant was instructed to read and sign a consent form approved by the Institutional Review Board. 
We conducted a five-minute training session to familiarize the participant with the driving simulator setup. 

The vehicle started driving in autopilot mode by default. Participants can decide whether to take over the vehicle to handle the incidents on the road.
If the participants did takeover, they were required to switch back to autopilot mode before arriving at the next intersection so that the vehicle can choose the next direction to go.
We asked the participants to periodically record their trust in the AV using the buttons on the steering wheel. 
It took about 40 minutes for each participant to complete the entire experiment. 

\subsection{Results} 

The computed maximal probabilities for satisfying the scLDTL specifications $\varphi_1$ and $\varphi_2$ are ${\rm Pr}_{{\mathcal{M}}}^{\max}(\varphi_1)=0.8357$ and ${\rm Pr}_{{\mathcal{M}}}^{\max}(\varphi_2)=0.7288$, respectively.

For each participant, we evaluate the satisfaction of each scLDTL specification $\varphi_i$ using the robot states $x$ and human trust levels $\Theta$ recorded in the experiment, and the induced trust beliefs $b_\Theta$. In total, 18 out of 21 ($18/21=0.8571>0.8357$) and 17 out of 21 ($17/21=0.8095>0.7288$) participants successfully complete the specifications $\varphi_1$ and $\varphi_2$ respectively, which validates the effectiveness of the proposed approach. Moreover, we further compare the trust-aware and trust-free routes for all drivers and both trails. The percentage of scLDTL satisfaction is 0.85 (trust-aware) vs 0.775 (trust-free). The average human trust is 4.6018 (trust-aware) vs 4.3933 (trust-free). One can see that the trust-aware policy outperforms the trust-free one. A video demonstration of the human experiment can be found at: \href{https://drive.google.com/file/d/1HLHBJe1Xn6Sa6BXVofYKTcKrpGjPeoSj/view?usp=drive_link}{link}.

We summarise three key observations gained from the experiments. First, observing the AV successfully handle the same incidents multiple times does not necessarily guarantee an increase in human trust. Second, if a human's trust remains consistently low for an extended period, it becomes challenging for them to re-establish trust in the AV. Third, having the capability to effectively address an incident beforehand can contribute to boosting human trust. Based on the feedback received after the experiment, participants have indicated that, had they noticed the car braking earlier in situations involving pedestrians, they might have considered the AV more trustworthy.

\section{Conclusions}

In this work, we presented a trust-aware motion planning approach for HRC. We demonstrated the suitability of scLDTL for describing the desired behaviours of HRC systems and an algorithm was proposed for solving the optimal policy synthesis problem. Human subject experiments were conducted on a driving simulator, validating the effectiveness of the proposed approach and providing valuable new insights. Additionally, we observed variations in trust dynamics among individuals, which will be further investigated in future research.

\bibliographystyle{IEEEtran}

\end{document}